\documentclass[journal]{IEEEtran}

\usepackage{graphics}
\usepackage{amsmath}
\usepackage{amssymb}
\usepackage{graphicx}
\usepackage{cite}
\usepackage{color}
\usepackage{algorithm}
\usepackage{algorithmic}

\newcommand{\etal}{\textit{et al}.}
\usepackage{url}
\usepackage{caption}
\usepackage{subfigure}
\begin{document}
%
\title{Reference-based Defect Detection Network}
%
%
%

\author{Zhaoyang~Zeng,~\IEEEmembership{Student Member,~IEEE,}
        Bei~Liu,~\IEEEmembership{Member,~IEEE,}
        Jianlong~Fu$^\dagger$,~\IEEEmembership{Member,~IEEE,}
        and~Hongyang~Chao$^\dagger$,~\IEEEmembership{Member,~IEEE}
\thanks{Z. Zeng and H. Chao are with the School of Computer Science and Engineering, Sun Yat-sen University. Email: zengzhy5@mail2.sysu.edu.cn, isschhy@mail.sysu.edu.cn}%
\thanks{Z. Zeng and H. Chao are with the Key Laboratory of Machine Intelligence and Advanced Computing (Sun Yat-sen University), Ministry of Education.}%
\thanks{B. Liu and J. Fu are with Microsoft Research Asia. Email: bei.liu@microsoft.com, jianf@microsoft.com}%
\thanks{$\dagger$ J. Fu and H. Chao are the corresponding authors.}%
\thanks{Digital Object Identifier 10.1109/TIP.2021.3096067}%
}

%
%

\markboth{Reference-based Defect Detection Network}%
{Shell \MakeLowercase{\textit{et al.}}: Bare Demo of IEEEtran.cls for IEEE Journals}
%




\maketitle


\begin{abstract}
The defect detection task can be regarded as a realistic scenario of object detection in the computer vision field and it is widely used in the industrial field. Directly applying vanilla object detector to defect detection task can achieve promising results, while there still exists challenging issues that have not been solved. The first issue is the texture shift which means a trained defect detector model will be easily affected by unseen texture, and the second issue is partial visual confusion which indicates that a partial defect box is visually similar with a complete box. To tackle these two problems, we propose a Reference-based Defect Detection Network (RDDN). Specifically, we introduce template reference and context reference to against those two problems, respectively. Template reference can reduce the texture shift from image, feature or region levels, and encourage the detectors to focus more on the defective area as a result. We can use either well-aligned template images or the outputs of a pseudo template generator as template references in this work, and they are jointly trained with detectors by the supervision of normal samples. To solve the partial visual confusion issue, we propose to leverage the carried context information of context reference, which is the concentric bigger box of each region proposal, to perform more accurate region classification and regression. Experiments on two defect detection datasets demonstrate the effectiveness of our proposed approach.
\end{abstract}


%

\begin{figure}
\centering
\includegraphics[width=0.8\linewidth]{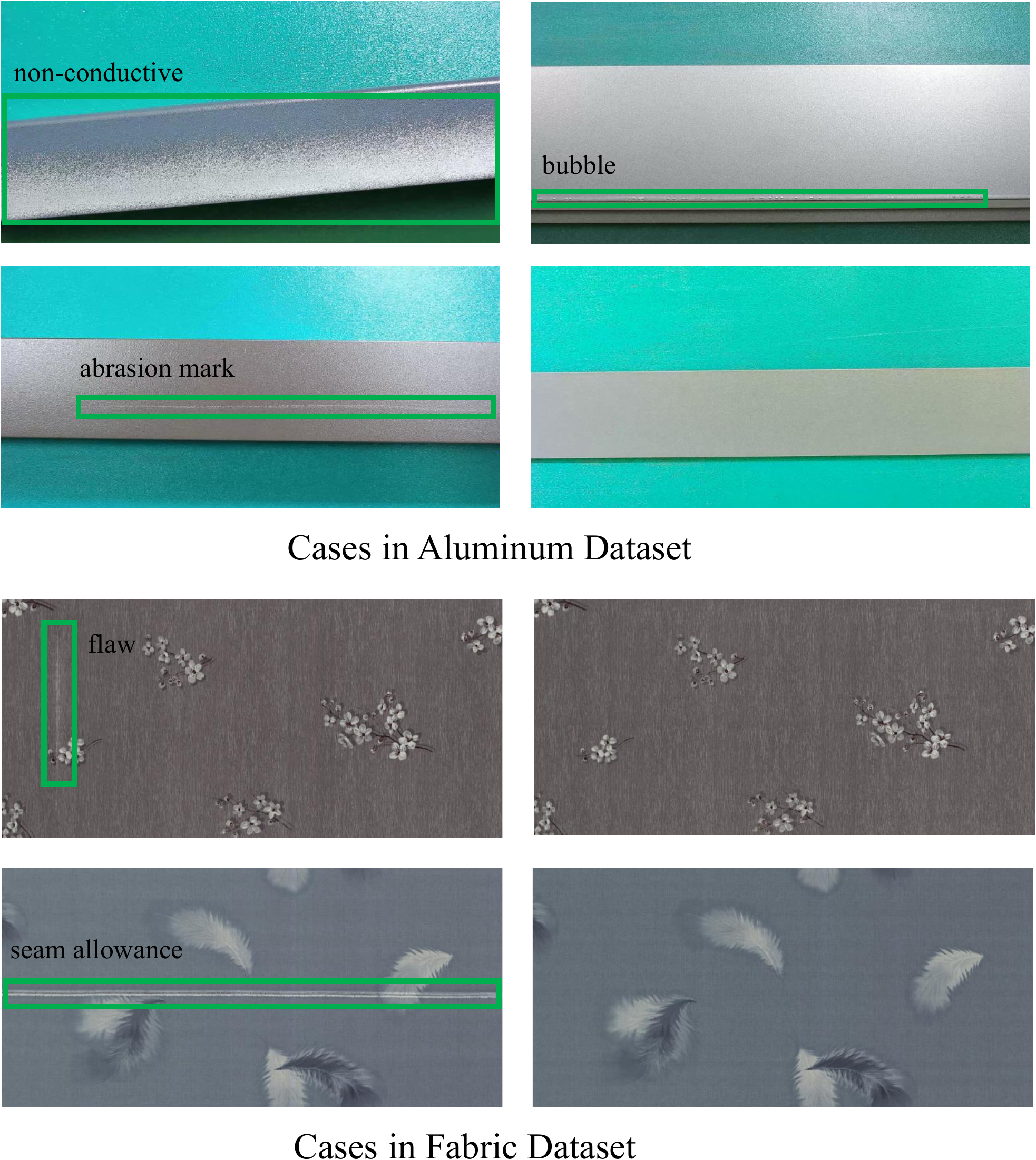}
\caption{Defect and normal samples of \textbf{Aluminum} and \textbf{Fabric} datasets. Green boxes indicate the defect annotations. The two defect-free samples given in \textbf{Farbic} dataset are well-aligned template images of the defective ones.}
\label{fig:samples}
\vspace{-4mm}
\end{figure}

\section{Introduction}
\IEEEPARstart{C}ONVOLUTIONAL neural networks (CNNs) have been widely adopted in computer vision tasks, including category classification \cite{he2016deep,szegedy2015going,huang2017densely,Fu_2019_NeurIPS}, instance detection/segmentation\cite{ren2015faster,redmon2016you,liu2016ssd,he2017mask,zeng2019wsod2}, semantic segmentation\cite{noh2015learning,ronneberger2015u,lin2019spatially} and cross-modality understanding\cite{huang2021seeing,su2019vl,liu2018beyond}. The excellent performance of CNNs credits to the powerful network architecture \cite{he2016deep,huang2017densely,szegedy2015going}, the effective mechanism design \cite{Fu_2017_CVPR,Fu_2019_TIP}, the development of hardware devices\cite{markidis2018nvidia}, and the large scale datasets with annotations \cite{deng2009imagenet,lin2014microsoft}.

In the industrial field, defect detection is an important step to guarantee product quality, which requires quality inspectors to recognize defective samples and find the defective area on the surface of industrial products such as fabric, and metal. Detecting defects by humans is laborious and time-costly because most defects are in thin or small shapes. Some researchers have noticed these pain points, and try to introduce CNNs to solve them in production environments \cite{wang2018fast,tong2017fabric,liu2018fabric,chen2017automatic}.

Since the possible defect needs to be accurately located in the defect detection task, it is intuitive to model the defect detection task as an object detection problem in the computer vision field. Early CNN-based approaches~\cite{wang2018fast,napoletano2018anomaly} to solve defect detection tasks most adopt a sliding-window algorithm and recognize each possible defect in the image patches. However, such strategies are inefficient and can not achieve the requirement of real application. Object detection is a wide-studied task, many approaches have been proposed to solve object detection tasks in more efficient and effective ways in these few years. Most recent object detection approaches can be well transferred to defect detection tasks without the need of specific designs. Some researchers have tried to applied successfully detectors, such as Faster R-CNN~\cite{ren2015faster}, SSD~\cite{liu2016ssd} and YOLO~\cite{redmon2016you}, to solve the defect detection task, and have achieved some promising results~\cite{cha2018autonomous,chen2017automatic}.

Although directly applying vanilla object detector to defect detection task can achieve promising results, there are still some unsolved challenging issues brought by defect detection task-specific characteristics. Firstly, the candidate images of defect detection are mostly taken from some specific cameras in the production pipeline, and thus the variance of background appearance is relatively small. The lack of background variance may mislead the network to over-fit the seen background, and thus limit the generation ability to unseen background patterns. For example, when the trained detector meets an image with an unseen background or texture pattern, it will be difficult to point out the defect area and will misclassify the unseen texture into defects. We call this problem as \textbf{texture shift (TS)}. Secondly, a partial defect box and a complete defect box may have a very similar appearance, as shown in the examples in Fig.\ref{fig:samples}, so it is difficult for the detector to distinguish partial defect box and complete defect box only based on their inside visual information. We call this issue \textbf{partial visual confusion (PVC)}. Existing object detection approaches still lack module to solve the above two issues.

In this paper, we focus on the two above issues and propose a Reference-based Defect Detection Network (RDDN) by involving two kinds of visual references. The first kind of reference is \textbf{template reference} (TR), which are samples that share the same global appearance and detailed texture with the candidate detected images. The unrelated texture information can be erased by subtracting the candidate images by their template reference in either image or feature levels. The second type of reference is \textbf{context reference} (CR), which is a concentric larger region than the given region of interest (ROI), and can provide additional contextual information for making category and boundary decisions.

We adopt two formulations for template reference. The first one is the well-aligned template images computed from the production machines, as examples shown in Fig.\ref{fig:samples}. The well-aligned template images and the candidate images are highly similar, and the difference only lies in the defective area. We propose three processing strategies to make full use of such well-aligned templates to solve the TS issue from three levels: image, feature, and region. The second one is pseudo template, which is generated by a \textbf{pseudo template generator} (PTG). The PTG can smooth the defective area, and transfer detective samples into the defect-free domain in the feature level. With the help of PTG, the defective area will be highlighted by a feature residual operation, and most background and texture information will be dismissed. The PTG can be integrated into vanilla object detectors, and can be trained by easy-achieved defect-free samples in an end-to-end manner.

The context reference indicates the concentric bigger box of the given box, which can provide additional information to solve the PVC issue. The PVC will confuse the vanilla detector on both category classification and boundary regression. Specifically, the partial defect boxes and the complete defect box may have similar features but will be assigned different category labels and bounding box regression targets during training. The key challenge is that the respective field of a box is most concentrated inside itself, and thus it is difficult to infer the endpoint of the defect if it only contains a part of the defect. We adopt the context reference by fusing the region features with their reference features and perform region-level category classification and bounding box regression on the fused features. Context reference can provide surrounding information to enlarge the difference of visual-similar partial and complete defect boxes in feature-level, and thus can benefit both category classification and boundary decision.

We implement the RDDN by integrating a template reference and context reference into classical two-stage detectors\cite{ren2015faster,cai2018cascade}. We evaluate the proposed RDDN on the \textbf{Aluminum} \cite{aluminum2018} and \textbf{Fabric} \cite{frabic2019} defect detection datasets, which are both released by Tianchi platform \cite{tianchi}. We make a detailed comparison to analyze the effectiveness of each component. The contribution of this paper can be summarized as follow:
\begin{itemize}
\item We analyze and summarize the challenging issues in defect detection tasks, especially the problems of existing common ways that directly apply object detection.
\item We propose to tackle these issues by a Reference-based Defect Detection Network (RDDN), which involves template references to erase texture variance and context references to provide additional contextual information.
\item We conduct extensive experiments on two defect detection datasets. The experimental results demonstrate the effectiveness of our approach.
\end{itemize}

\section{Related Works}
\subsection{Object Detection}
Considering the industrial requirement and annotating cost, we treat the defect detection task as an object detection problem with box-level annotations, where the ``object'' is some defects of interest. Currently, the most accurate object detectors most follows the two-stage pipeline, which first generate region proposals on images, then perform classification and regression on each proposal. The first two-stage detector is R-CNN\cite{girshick2014rich}, which is proposed in 2014. R-CNN generates region proposals by selective search\cite{uijlings2013selective}, and uses an SVM classifier to predict the category of each region proposal. Later on, Fast R-CNN\cite{girshick2015fast} introduces ROI-Pooling layer to make the feature extraction process of CNN backbone shared so that greatly reduces the inference time. Faster R-CNN\cite{ren2015faster} proposes region proposal network (RPN) to dynamically generate region proposals from feature maps, which leads the wave of the latter two-stage detectors. A great number of approaches are proposed based on the architecture of Faster R-CNN in these few years. Lin \etal\cite{lin2017feature} propose feature pyramid networks, which involve feature fusion from different semantic levels to enhance the representation ability, and dynamic map region proposals of different sizes to different pyramid levels, so that each proposal can be assigned a better mapping resolution. FPN is now be considered as a default basic setting in recent two-stage detectors. Cai \etal\cite{cai2018cascade} find that only adjust the region proposal once still can not achieve a good boundary precision, and thus propose cascade R-CNN to progressive refine each region proposal several times, and so that the localization accuracy can be greatly improved.

In this paper, we focus on solving the challenging issues for defect detection. We follow the pipeline of two-stage detectors to design our solution, and the idea can be transferred to other detection approaches.

\subsection{Defect Detection}
There  have  been  already  some  works  that  try  to  explore convolutional neural networks into industrial defect detection. Wang \etal \cite{wang2018fast} proposes to adopt CNNs to recognize the defect type in a given patch and use a sliding window to detect the defect positions in images. It can achieve good detection accuracy, while the lacking of computing efficiency will limit the applicability in the wild. Napoletano \etal \cite{napoletano2018anomaly} use CNN pre-trained on ImageNet dataset to extract features from image patches and form a dictionary, and detect defect by computing the similarity between given patch feature and features in a dictionary. Tao \etal \cite{tao2018automatic} focus on metallic defect detection, and propose a coarse-to-fine detection pipeline by first coarsely segments the defect area, then applying fine defect recognition. Cha \etal \cite{cha2018autonomous} apply Faster R-CNN on defect detection field and survey the influence of backbone network architecture. Chen \etal \cite{chen2017automatic} proposes a three-stage framework to detect defects on the catenary support device from coarse to fine. SSD\cite{liu2016ssd} and YOLO\cite{redmon2016you} detectors are used in the first two stages, and a CNN classification network is adopted to judge if the detected area contains defects.

The TS issue has been studied in previous works. Tong \etal \cite{tong2017fabric} proposes a non-locally centralized sparse representation (NCSR) to reconstruct defect-free images and perform defect detection based on the residual images of reconstructed images and input images. Such an algorithm is suitable for simple cases but might be failed when meeting complex textures. Liu \etal \cite{liu2019multistage} proposes a multi-stage generative adversative network to generate defective training samples based on defect-free images. This is essentially a kind of data extension strategy, which does not fundamentally overcome the problem.

\section{Basic Settings}

\subsection{Datasets}

We first introduce two defect detection datasets we used in this work. The two datasets we used are aluminum defect \cite{aluminum2018} and fabric defect \cite{frabic2019}, which are both released by Tianchi platform \cite{tianchi}. We will use \textbf{Aluminum} and \textbf{Fabric} to indicate these two datasets for short for the rest of this paper. The cases in Fig.\ref{fig:samples} all come from these two datasets. The detailed statistics of these two datasets are as follows:
\begin{itemize}
    \item \textbf{Aluminum} consists of $4356$ images and has $10$ kinds of defect categories. Among these, $1351$ images are normal samples, which do not contain any defect. We randomly select $500$ images for testing, and the rest images are all used for training. Besides, there are also normal samples in the testing split.
    \item \textbf{Fabric} consists of $3522$ images belonging to $15$ defect categories, where $416$ images are normal samples. It is worth noting that all images in this dataset are paired with a well-aligned template.
\end{itemize}

\begin{table}[t]

    \centering
    \begin{tabular}{c|c|c|c}
         defect name & \#boxes & minimum area & maximum area\\
         \hline
         non-conductive & 392 & 3,956 & 2,398,720 \\ 
         flow mark & 100 & 16,133 & 1,078,227 \\ 
         abrasion mark & 190 & 2,196 & 2,944,258 \\ 
         color mismatch & 307 & 632,320 & 3,837,440 \\ 
         orange peel & 172 & 151,040 & 3,274,240 \\ 
         bubble & 194 & 336 & 472,120 \\ 
         leakage & 577 & 4,012 & 2,373,120 \\ 
         dirty spot & 689 & 806 & 4,042,240 \\ 
         corner leakage & 378 & 3,843 & 931,840 \\ 
         pit & 372 & 66,560 & 1,948,160\\ 
         \hline
    \end{tabular}
    \caption{Data analysis of \textbf{Aluminum} dataset.}
    \label{tab:aluminum_analysis}
\vspace{-3mm}
\end{table}

\begin{table}[t]

    \centering
    \begin{tabular}{c|c|c|c}
         defect name & \#boxes & minimum area & maximum area\\
         \hline
stained & 544 & 2,618 & 6,901,653 \\
broken figures & 381 & 27,648 & 6,942,796 \\
water mark & 4,135 & 2 & 6,926,099 \\
knots & 218 & 1,516 & 6,946,816 \\
seam allowance & 437 & 265 & 112,727 \\
joining stencil mark & 1,074 & 18 & 411,978 \\
insect & 2,426 & 14 & 6,946,816 \\
broken hole & 104 & 1,046 & 6,946,816 \\
pleats & 307 & 195 & 6,930,866 \\
flaw & 256 & 49 & 1,733,395 \\
misprinting & 319 & 862 & 1,196,417 \\
wax spot & 136 & 88 & 6,627,136 \\
color shade & 3,187 & 20 & 4,216,730 \\
crease mark & 106 & 121 & 6,928,369 \\
others & 134 & 582 & 9,396 \\
         \hline
    \end{tabular}
    \caption{Data analysis of \textbf{Fabric} dataset.}
    \label{tab:fabric_analysis}
\vspace{-3mm}
\end{table}
The detailed analysis of each defect category of the two datasets can be found in Table~\ref{tab:aluminum_analysis} and Table~\ref{tab:fabric_analysis}. For both datasets, we randomly select $500$ images for testing and use all the rest images for training. We use mAP@0.5 as a metric to evaluate the performance of both datasets.

\subsection{Baseline Detector}
We introduce a baseline to solve the defect detection tasks, which is built upon the widely used architecture Faster R-CNN. We first give a brief introduction of the Faster R-CNN, then we will introduce the setting of each component in detail.

\begin{figure}[t]
    \centering
    \includegraphics[width=\linewidth]{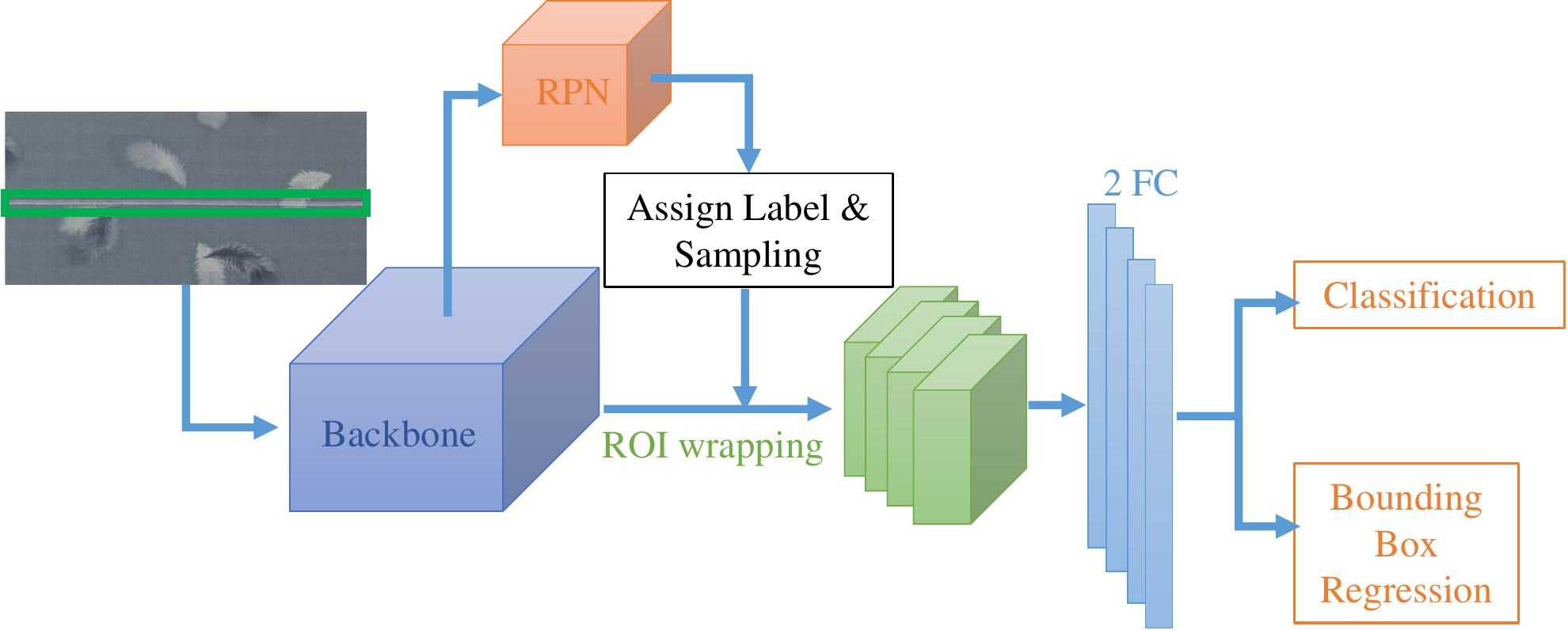}
    \caption{The overview of Faster R-CNN. The Faster R-CNN is constructed of a CNN backbone for image feature extraction, an RPN for region proposals generation, an ROI wrapping layer for fix-sized region feature extraction and an ROI head for regions classification and regression.}
    \label{fig:framework}
\end{figure}

Faster R-CNN is first proposed in \cite{ren2015faster}, which is the basic architecture of most recent two-stage detectors. As the overview shown in Fig.\ref{fig:framework}, the Faster R-CNN consists of a CNN backbone, an RPN, an ROI wrapping layer, and an ROI head. We will describe the details of all components as follows.

\subsubsection{CNN Backbone}
The CNN backbone is used to extract image features. To construct a fair comparison, we adopt ResNet-50 and ResNet-101 two kinds of CNN backbones in our later experiment settings. We equip the CNN backbone with a feature pyramid network (FPN), which is first proposed in \cite{lin2017feature} and is adopted as a default setting in many later works because of its effectiveness. FPN can enhance the feature representation ability of low-level features, as well as to enlarge the mapping resolution of small boxes. Such an advantage can benefit defect detection especially for those defect boxes with thin or small shapes. Given the input image $I$, we denote the output feature of the second stage to the fifth stage of ResNet as 
\begin{equation}
    \begin{aligned}
    \textbf{C}^I=\{C^I_2, C^I_3, C^I_4, C^I_5\}.
    \end{aligned}
\end{equation}
We denote the output of FPN as
\begin{equation}
    \begin{aligned}
        \textbf{P}^I=\{P^I_2, P^I_3, P^I_4, P^I_5\},
    \end{aligned}
\end{equation}
which is computed as
\begin{equation}
\begin{aligned}
L_5 =& conv_{1\times 1}(C^I_5), \\
L_i =& conv_{1\times 1}(C^I_i) \\
&+ upsample(L_{i+1})  (2\leq i \leq 4), \\
P^I_i =& conv_{3\times 3}(L_i)  (2\leq i \leq 5),
\end{aligned}
\end{equation}
where $conv_{1\times 1}$ denotes $1 \times 1$ convolutional operation, $conv_{3\times 3}$ denotes $3 \times 3$ convolutional operation, and $upsample$ denotes a 2 times bilinear upsampling operation. We follow \cite{lin2017feature} to make all $conv_{3\times 3}$ blocks share weights. In our experiments, the output channel of FPN is set to $256$.

Besides, we also try to equip the ResNet with deformable convolution \cite{dai2017deformable, zhu2019deformable} to evaluate its effectiveness on defect detection tasks. We follow \cite{zhu2019deformable} to replace all $3\times 3$ convolutional layers in the third, fourth, and fifth stages in ResNet by deformable convolutional layers. The backbones with deformable convolution layers will be denoted as ResNet-50-DCN and ResNet-101-DCN in our experiments.

\subsubsection{RPN}
The region proposal network will generate region proposals for the features from all levels of $\textbf{P}^I$ based on a serial of pre-defined anchors. We use the default anchor setting used for COCO dataset. The anchor stride is set to $8$, and the aspect ratios are set to $\left[1:1, 1:2, 2:1\right]$. Such anchor setting will be applied on every pyramid level in FPN. Therefore, the real anchor sizes are $\left[32, 64, 128, 256\right]$. RPN will produce an objectness score and a regression parameter for each anchor. After threshold filtering and NMS, the region proposals with high objectness scores will be selected. We denote the set generated region proposals as
\begin{equation}
    \begin{aligned}
    R^I=\{r^I_1, r^I_2, \cdots, r^I_n\},
    \end{aligned}
\end{equation}
where $n$ is the number of output region proposals.

\subsubsection{ROI wrapping Layer}
We follow \cite{he2017mask} to adopt ROI-Align as the ROI wrapping layer. Comparing with ROI-Pooling proposed in \cite{girshick2015fast}, ROI-Align uses bilinear interpolation to computer the feature value in float position, which can reduce the precision missing in the floor operation.

The ROI wrapping layer is used to extract a fix-sized feature for each region proposal, based on the FPN features $\textbf{P}^I$. We follow \cite{lin2017feature} to map region proposals to pyramid levels according to their sizes for feature extraction. Specifically, given a region proposal with width $w$ and height $h$, the mapping pyramid level is calculated by $\lfloor log_2(\frac{\sqrt{wh}}{56}\rfloor$,
and further clipped into $\left[2, 5\right]$. The output size of the ROI Align layer is set to $7 \times 7$.

\subsubsection{ROI Head}
We adopt the widely used two-MLP ROI head to perform region-level classification and regression. After obtaining the ROI features, we first break the spatial dimension and flat each feature into a feature vector. We then use two $1024-$d fully-connected layers followed by the ReLU activation function to embed each feature vector. For region $r^I_i$, we denote the output of the second fully-connected layers after activation as $f^I_i$ and call it region feature in the rest of this paper. We follow \cite{ren2015faster} to apply a $c+1$ categories classification on $F$, and adopt a class-specific bounding box regression module.

We also involve Cascade R-CNN architecture~\cite{cai2018cascade}, where a total of three ROI heads are built. All three ROI heads have the same setting. Region proposals will be refined by all ROI heads one by one in both training and testing stages.

\subsection{Training Strategies}
There are tens of hyperparameters inside the framework of Faster R-CNN. We empirically set their values. Most values are borrowed from open-source codes~\cite{chen2019mmdetection}, which are proved to be effective on MS-COCO benchmark, and the other values (e.g. input size, ROI head IOU thresholds) are decided by local experiments. We use these values to construct a strong baseline detector, and will not modify them in later experiments.

\subsubsection{Input size}
We adopt multi-scale training since it is proved to be effective in many works \cite{chen2019mmdetection,chen2019hybrid}. We only use single-scale testing because it is more acceptable in the industrial application scenario. Given a training input image, we first randomly select a size in the range $[600, 1000]$, then resize the short edge of the input image to such size, and limit the long edge to $1600$. For testing, we resize the short edge of the input image to $800$, and limit the long edge to $1600$.

\subsubsection{Pre-trained parameters}
We adopt the models trained on MS-COCO dataset \cite{lin2014microsoft} to initialize the network. The pre-trained models are publicly accessible. Ton initialize the network, we abandon the parameters of the classification and regression layers in both RPN and ROI head because of the misalignment of the parameter dimension. Since the image number of the two datasets is relatively small, using COCO pre-trained models can increase both convergence speed and final performance. Besides, we follow \cite{chen2019mmdetection} to freeze all parameters in the stem and the first stage of ResNet during training.

\subsubsection{Label Assignment and Sampling}
When training RPN, anchors who have IOU larger than $0.7$ with any ground truth box will be considered as positive anchors, and have IOU less than $0.1$ with all ground-truth boxes will be considered as negative. All other anchors will not contribute to the training. When RPN generates region proposals, we apply NMS with a threshold of $0.7$ on the top $20000$ high-confident ones and keep $2000$ region proposals finally. We assign a category label to each region proposal according to the Max-IOU strategy. When using Faster R-CNN architecture, a region proposal is utilized as a positive proposal if it has IOU larger than $0.5$ with any ground truth boxes, otherwise, it will be considered as a negative one. For Cascade R-CNN, the IOU thresholds will be set to $0.4$, $0.5$, $0.6$ for the three ROI heads, respectively. Comparing with the origin setting proposed in \cite{cai2018cascade}, we decrease the IOU thresholds for all heads to involve more positive proposals for training. Besides, all ground-truth boxes will be considered as positive region proposals.

We randomly select $128$ positive samples and $374$ negative samples to form an ROI batch. We will pad the batch size to $512$ by negative samples.

\subsubsection{Loss}
We apply binary classification and smooth-L1 loss for RPN and apply category-wise softmax classification loss and class-specific smooth-L1 bounding box regression loss for ROI heads. When performing region classification, we add a ``background'' category to indicate negative samples. We refer to \cite{ren2015faster} for the detail formulation of the loss functions.

\subsubsection{Optimizer and learning rate}
We train all models on 8 NVIDIA V100 GPUs with a batch size of $1$ per GPU. We adopt SGD as the optimizer, and set the initial learning rate to $0.01$. We train models for $24$ epochs, and decay the learning rate by $10$ times at $16^{th}$ and $22^{th}$ epochs. We adopt a linear warmup strategy at the first $500$ training iterations.

\subsubsection{Post processing}
When we obtain the classification and bounding box predictions of each region proposal, we apply NMS with a threshold $0.5$ for each category, and filter out those boxes whose classification score lower than $0.01$. We set the maximum detection number of each image to $100$ based on the confidence of each box.

\section{Approach}
In this section, we will detail introduce the data characteristics of defect detection tasks and analyze their brought texture shift (TS) and partial visual confusion (PVC) issues. To solve the issues, we proposed a reference-based defection detection network (RDDN) by involving two kinds of reference, including \textbf{template reference} and \textbf{context reference}, to improve the defect detector and boost the detection performance. The RDDN is built upon a standard Faster R-CNN detector, and can be easily extended to other detection networks.

\begin{figure*}[t]
\centering
\includegraphics[width=0.8\linewidth]{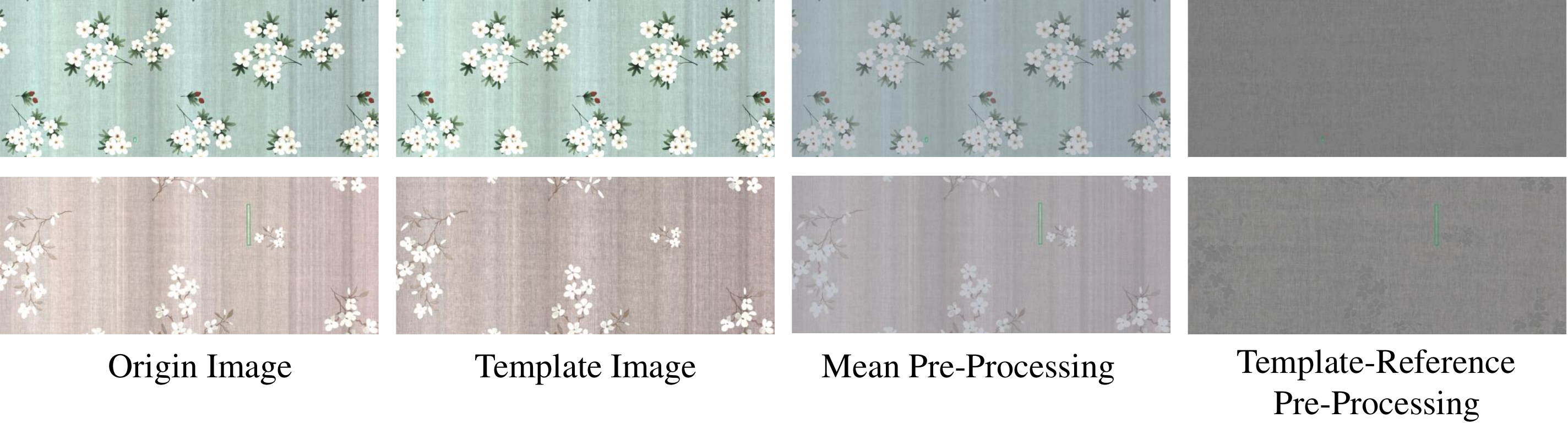}
\caption{Examples of different image pre-processing strategies. The first two columns show origin images and corresponding template images, respectively. The last two columns are
mean pre-processing images and template reference pre-processing images, respectively. All images in the last two columns are normalized to $\left[0, 255\right]$ to avoid negative pixel value for visualization. Boxes in green indicate annotated ground truths. [Best viewed in color]}
\label{fig:tgp}
\end{figure*}

\subsection{Template Reference}

Given a candidate image $I$, we decompose it into
\begin{equation}
\begin{aligned}
I = I_T + \delta,
\end{aligned}
\end{equation}
where $T$ indicates the texture of $I$, which stores the expected appearance information, and $\delta$ is the color shift. In most cases, $|I_T|\gg |\delta|$, since ideally $\delta$ should be in a zero-mean distribution. We consider that the defect information will be all contained in $\delta$. If a model (or part of a model) wants to learn the defect knowledge of $\delta$ based on $I$, $I_T$ will disturb the training progress by guiding the model to fit its data distribution. In the testing stage, if an image with an unseen texture is provided, the trained detector may give the wrong prediction because the given image template is out of the training distribution. We call this phenomenon \textbf{texture shift (TS)}. This issue can be solved by training the defect detector based on $I-I_T$, because the interference information will be dismissed in this formulation. We propose to use template reference to approach $I_T$, and reduce the negative impacts brought by TS.

In some specific application scenarios, we have a chance to achieve well-aligned templates for defect samples. For example in Fig.\ref{fig:samples}, in \textbf{Fabric} dataset, each image is paired with a template image, which is obtained from the production pipeline. Such templates contain all texture information of candidate samples, serving as a reference to help improve defect detection performance. We propose three processing strategies to make full use of the template, including a pre-processing, an inter-processing, and a post-processing strategies.

\subsubsection{Template Reference Pre-processing}

Before feeding images into CNN backbone, a common operation is to normalize images by pre-computed mean and var. Specifically, given the input image $I$, the normalized image is computed by
\begin{equation}
\begin{aligned}
I' = \frac{I - mean}{var},
\end{aligned}
\end{equation}
where $mean$ and $var$ are pre-computed values. Typically, we adopt the pixel mean value and standard deviation of each channel computed on ImageNet \cite{deng2009imagenet} dataset. The target of normalization operation is to reduce the value range of computed feature maps, and thus the gradient backwarded by loss function will be more stable.

In the defect detect detection area, especially in \textbf{Fabric} dataset, the candidate detected images may have different kinds of textures, as shown in the first column of Fig.\ref{fig:tgp}. These textures may bring difficulties to network learning. Besides, if using a fabric image whose texture is never seen in the training stage, it would be difficult for the detector to detect. This issue can be solved with the help of template images. We propose a \textbf{template reference pre-processing (TR-Pre)} strategy to eliminate the difference in textures. Specifically, we replace the pre-processing mean value with the corresponding template image. The modified pre-processing is formulated as
\begin{equation}
\begin{aligned}
I' = \frac{I - T}{2var},
\end{aligned}
\end{equation}
where $T$ denotes the corresponding image template. Note that in the standard pre-processing, the value range of each pixel is $[0,255]$, so the value range of $I-mean$ is $[-127.5, 127.5]$ if we assume $mean=127.5$, while the value range of $I-T$ is $[-255, 255]$. We divide the value by $2$ to better match the network pre-trained parameters since the first few convolutional layers are fixed. We still adopt the $var$ value as the same in the standard pre-processing to make the input satisfy pre-trained parameters as more as possible.

In the original mean pre-processing progress, although a normalization was applied, the visual difference of samples still can not be dismissed. If template reference pre-processing is applied, the visual difference can be erased a lot, so that the network does not need to overcome such variety and focus more on defect detection. We compute the pixel standard deviation on \textbf{Farbic} dataset to evaluate the diversity of processed images. When adopting mean pre-processing, the standard deviation is $\left[56.61, 55.42, 51.23\right]$ for R, G, B channels, respectively. The standard deviation will reduce to $\left[20.05, 20.59, 20.15\right]$ using template reference pre-processing. Template reference pre-processing can reduce visual diversity, while still keep the visual information of defects, thus can benefit the network training.

\subsubsection{Template Reference Inter-processing}
\begin{figure}[t]
\centering
\includegraphics[width=\linewidth]{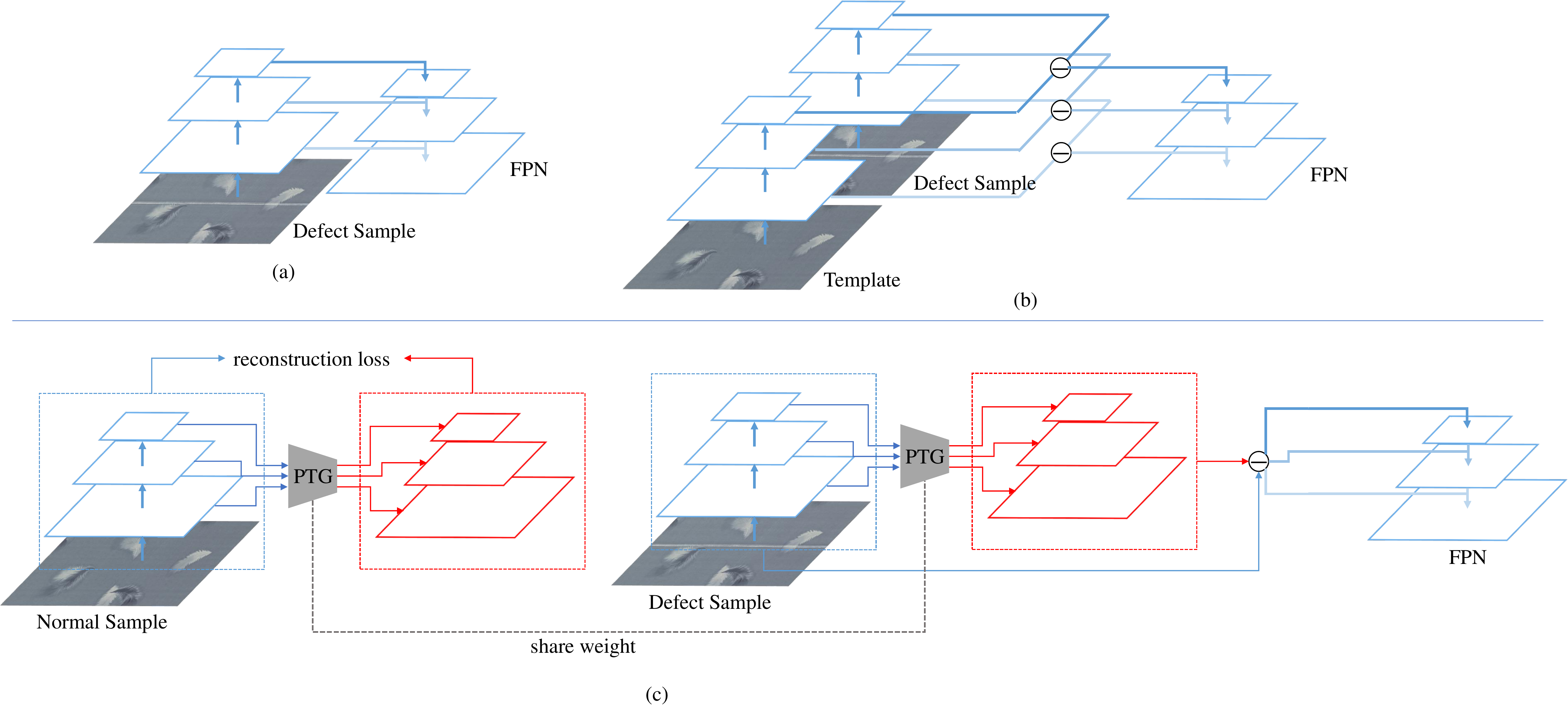}
\caption{(a) Overview of FPN; (b) Overview of template reference inter-processing, where the FPN is built upon the subtracting of the defect and template features; (c) Overview of template reference inter-processing with pseudo template generator (PTG).}
\label{fig:tginter}
\vspace{-4mm}
\end{figure}
Although template images and candidate detected images are aligned, there still exists a few-pixel gap. The pixel gap will bring noise edge when applying minus operation in some pattern edge area. The noise edge caused by the pixel gap looks similar to a defect and thus may confuse the detector.

\begin{figure}[t]
\centering
\includegraphics[width=\linewidth]{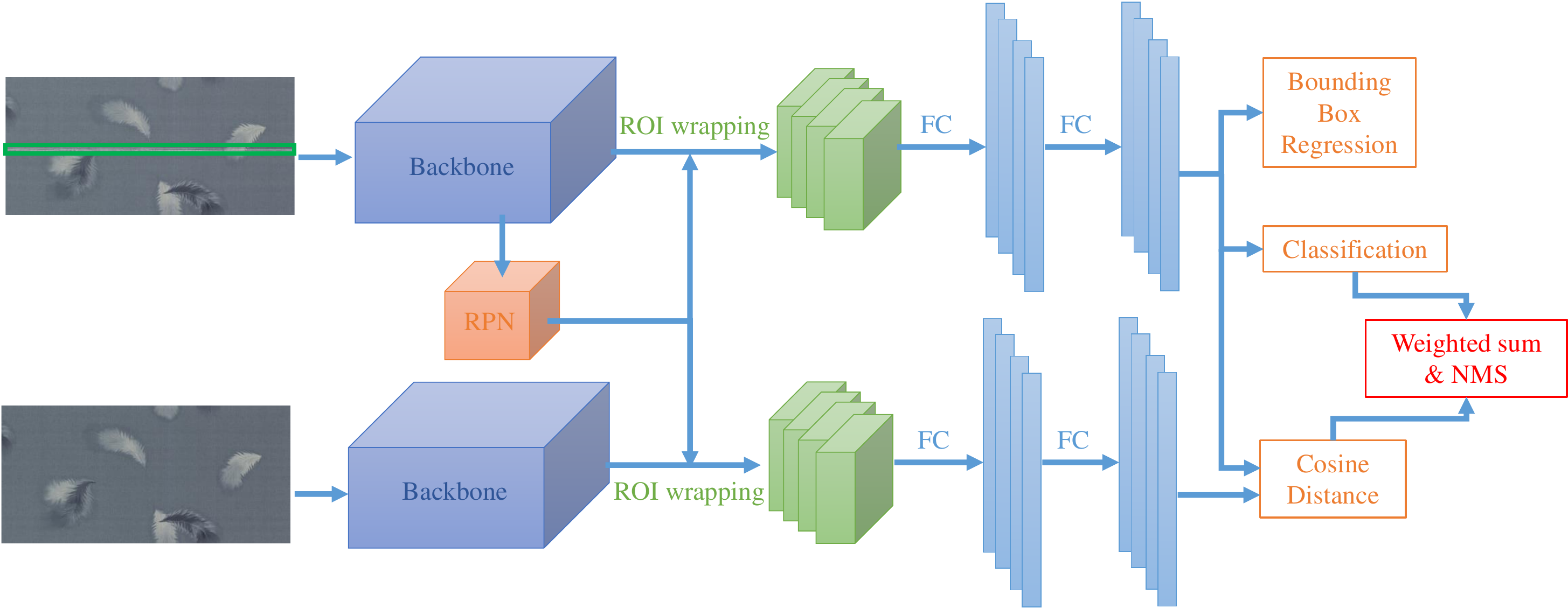}
\caption{Pipeline of template reference post-processing. In the inference, the candidate detected image, as well as its corresponding template image, will be fed into the backbone to extract image features. RPN will only generate region proposals for candidate images. We extract two kinds of region features from both image features for each region proposal. The cosine distance of two kinds of region features will be combined with the classification score to perform NMS.}
\label{fig:tgpost}
\vspace{-4mm}
\end{figure}

The gap between template and defect images exists at the pixel level, while will be smoothed in feature-level because of the large receptive field of each feature map position of CNN, while the response of defect will exist in both pixel-level and feature-level. Therefore, it is intuitive to dismiss the influence of pixel-gap in feature-level. Inspired by this, we propose \textbf{template reference inter-processing (TR-Inter)} to leverage the template to guide the network training at the feature level. The details of TR-Inter are described as followed. Given the input image $I$ and the template image $T$, we first feed them into the ResNet backbone and achieve the feature from all stages. We denote the ResNet feature of $T$ as
\begin{equation}
\begin{aligned}
C^{T}=\{C^{T}_2, C^{T}_3, C^{T}_4, C^{T}_5\}.
\end{aligned}
\end{equation}
We apply the minus operation on the output of all ResNet stages, then build FPN based on the minus features. The modified FPN feature is computed by

\begin{equation}
\begin{aligned}
L_5 =& conv_{1\times 1}(C^I_5 - C^T_5), \\
L_i =& conv_{1\times 1}(C^I_i - C^T_i) \\
&+ upsample(L_{i+1}) (2\leq i \leq 4), \\
P^I_i =& conv_{3\times 3}(L_i) (2\leq i \leq 5),
\end{aligned}
\label{eqn:trinter}
\end{equation}

All the rest operations are kept the same as the baseline defect detector. Since template reference inter-processing applies minus operation in the feature level, the pixel gap will be smoothed by the convolutional and pooling operations. We apply the minus operation on the output of all ResNet stages to maximize the use of ImageNet pre-trained parameters.

\subsubsection{Template reference Post-processing}

Although template images can be easily achieved, in some application scenarios template images may be missing. For example, if the candidate detected image is not taken from the specific position, then template images are not available, and thus the defect detection model trained with template reference pre-processing or inter-processing is not usable.

We propose a new algorithm to utilize the templates to re-score the predicted boxes, called \textbf{template reference post-processing (TR-Post)}. Different from TR-Pre and TR-Inter, TR-Post does not need to train another detector and can be directly plugged into the baseline detectors.

We propose template-guided post-processing to leverage region feature similarity to re-score each region proposal. Specifically, the classification head can produce score $P^I$, whose dimension is $n \times (c+1)$, where $n$ and $c$ denote the number of region proposals and categories, respectively. After obtaining region proposals $R^I$, we will use all region proposals to extract fully-connected region features on both defect image $I$ and template $T$. The region features extracted on $I$ and $T$ are denoted as $F^I$ and $F^T$, respectively.

The similarity of $F^I$ and $F^T$ can evaluate how likely the regions belong to a defect. We adopt cosine distance as the similarity evaluation metric. We use
\begin{equation}
    \begin{aligned}
    P^I\cdot (1-cos<F^I, F^T>)
    \end{aligned}
\end{equation}
as the final score of regions, where
\begin{equation}
\begin{aligned}
cos<F^I, F^T> = \frac{F^I \cdot F^T}{\left\|F^I\right\|\left\|F^T\right\|}.
\end{aligned}
\end{equation}
Since $F^I$ and $F^T$ are extracted from a ReLU layer, so the value range of $cos<F^I, F^T>$ is $[0,1]$. Such scores take both classification confidence and the similarity with template into consideration, and thus have richer representation ability. If the cosine distance of a region proposal has a large value, which means such a region looks similar to the template, thus it should have a low score. And oppositely, if the cosine distance has a small value, which means it has a large visual difference with the template, thus it should have a high score.

TR-Post can be viewed as a plug-in post-processing module, who has strong flexibility. When performing prediction, if templates are available, then we can apply TR-Post to obtain more accurate detection. If templates are not available, we still can ignore such a post-processing module, and only use the trained detector for prediction.

\subsection{Pseudo Template Generator}
The template reference requires well-aligned templates as guidance, which may not always available in most real application scenarios. However, normal samples are much easier to achieve when they are compared with well-aligned templates. Motivated by the success of generation networks \cite{goodfellow2014generative}, we propose to integrate a generation network to generate pseudo template feature based on defect samples, and leverage the normal samples as supervision. Since the input image of the detectors is in relatively high resolution (i.e. $800\times 1333$), it is difficult to directly generate such high-resolution pseudo templates, we plugin the generator in feature level, and apply template reference inter-processing to reduce the texture variance.

We call the proposed module as \textbf{pseudo template generator (PTG)}, whose overview can be found in Fig.\ref{fig:tginter}. In each iteration, we will randomly select a defect sample $I$ and a normal sample $N$ as the input. We first feed the two samples into the CNN backbone, and achieve their ResNet feature $\textbf{C}^I$ and $\textbf{C}^N$, where
\begin{equation}
    \begin{aligned}
    \textbf{C}^N=\{C^N_2, C^N_3, C^N_4, C^N_5\}.
    \end{aligned}
\end{equation}
We involve a generator for each stage, where the generators all have encoder-decoder architectures. We denote the generator of the $i^{th}$ stage as $G_i$. Each $G_i$ consists of three convolutional layers and three de-convolutional layers. The strides of the last two convolutional layers and the first two de-convolutional layers are set to $2$. We equip all layers in $G_i$ except the last de-convolutional layer with ReLU activation layer. All hidden dimensions are set to $256$, and the output dimension of $G_i$ is set as the same as its input dimension. We adopt MSE reconstruction loss on normal samples to enable the generator to generate pseudo normal samples, which is denoted by

\begin{equation}
    \begin{aligned}
        Loss_g = \sum_{i=2}^{5}||C^N_i - G_i(C^N_i)||_2
    \end{aligned}
\end{equation}

When integrating the PTG into the Faster R-CNN, Eqn.~\ref{eqn:trinter} is modified as
\begin{equation}
\begin{aligned}
L_5 =& conv_{1\times 1}(C^I_5 - G_5(C^I_5)), \\
L_i =& conv_{1\times 1}(C^I_i - G_i(C^I_i)) \\
&+ upsample(L_{i+1}) (2\leq i \leq 4), \\
P^I_i =& conv_{3\times 3}(L_i) (2\leq i \leq 5).
\end{aligned}
\label{eqn:trinter2}
\end{equation}

The PTG and the Faster R-CNN can be end-to-end joint-trainable. Since the network is joint-trained, the detection losses can prevent the $G$ from coverage to an identity mapping.

\begin{figure}[t]
\centering
\includegraphics[width=\linewidth]{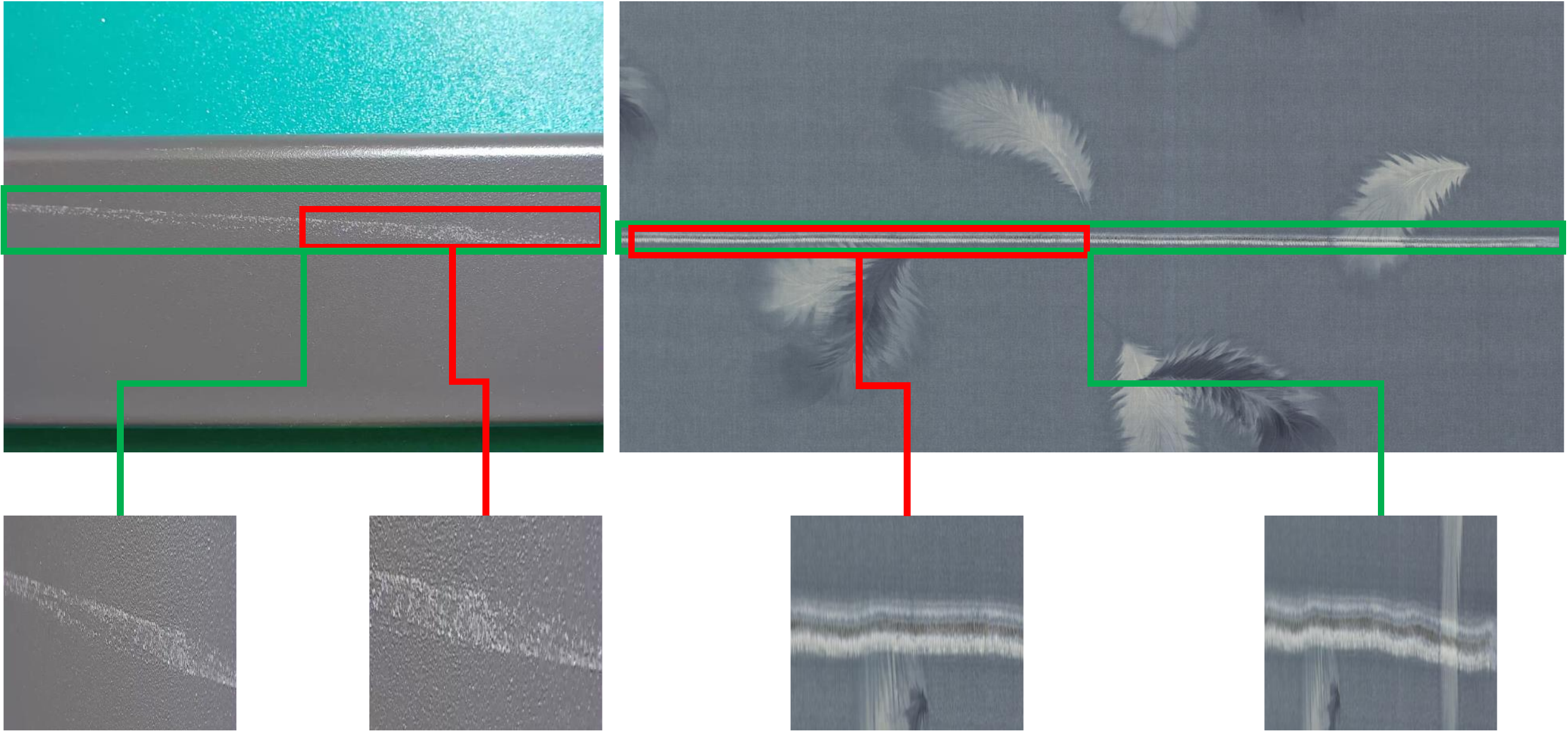}
\caption{Defect cases in \textbf{
Aluminum} and \textbf{Fabric} datasets. Green boxes indicate positive predictions, and red boxes are false negatives. When wrapping the boxes into the same size, the visual appearance of positive and negative boxes are very similar, and will bring difficulty to network learning.}
\label{fig:pvc}
\vspace{-3mm}
\end{figure}

\subsection{Context Reference}
\label{sec:pvc}

In the classical object detection task, a complete object and an object part usually have a large visual difference, thus it is relatively easy for the detector to classify them. However, in the defect detection field, there may be a high visual similarity between a complete defect and a defect part. We call this phenomenon \textbf{partial visual confusion (PVC)}. For example in Fig.\ref{fig:pvc}, the complete defect boxes in green look very similar to the defect part boxes in red, while the detecting target is only the complete defect.

PVC bring confusion to the defect detector from the following two aspects. First, when training the region classifier, a complete defect box will be assigned a specific category label, while a partial defect box will be assigned as background. The two boxes have a similar visual appearance, and thus will have a similar visual feature. Using a similar visual feature to fit different targets will bring difficulties to network training. Second, from human perception, given a larger proposal, we can easily fine-adjust it only according to the visual information inside, while it is difficult to adjust a smaller proposal to the correct location. The defect detector also has the same problem. In the design of Faster R-CNN, the ROI-wrapping layer is applied to wrap region features into a fix-size. When performing wrapping, the feature of each target position is sampled or computed from a few specific positions inside the region proposal box. Although CNN will enlarge the receptive field of each feature pixel, as discussed in \cite{luo2016understanding}, the surrounding area only provides little additional perception information, which may be not enough for the network to infer the box boundary.

To solve the two issues caused by PVC, we propose \textbf{context reference (CR)} to utilize the context information to help distinguish partial defects and decide defect boundary. For easier representation, each given region proposal $r$ can be represented as a quadruple denoted as $r=(x,y,w,h)$, where $x$ and $y$ are the coordinate center position, and $w$ and $h$ are the specific width and height receptively. We denote the contextual region of $r$ as $r_c=(x,y,kw,kh)$, where $k$ is a pre-defined coefficient and satisfies $k>1$. From the design of context region, we can find that $r_c$ has the same center position as $r$, while has a larger region size. We will clip $r_c$ to make sure that it will not exceed the border of the input image. In our experiments, we find that $k=1.5$ performs the best.

We follow the above method to compute the context region proposals of $R^I$ and denote whose region features as $F^I_c$. We concatenate $F^I$ and $F^I_c$ according to channel dimension and use the concatenated feature to perform category classification and bounding box regression.

Utilizing context information has already been proposed in \cite{he2016deep}. The difference between \cite{he2016deep} and our proposed method is that they consider the whole image as the context box, while we use the concentric larger box as a context box. Using the whole image can use the environment information to help detect objects while using the surrounding context focus more on region-level, which is more suitable for defect detection.

\section{Experiments}
We conduct experiments on \textbf{Aluminum} and \textbf{Fabric} datasets to evaluate the effectiveness of the proposed reference-based fabric defect detector. We firstly evaluate the performance of the baseline models on both datasets and then based on the baseline we conduct ablation experiments to evaluate the effectiveness of the two proposed reference modules. Finally, we evaluate the full setting of our proposed reference-based defect detection network.

\begin{table}[t]
\small
    \centering
    \begin{tabular}{c|c|c|c}
    \hline
        Detector & Backbone & \textbf{Aluminum} & \textbf{Fabric}\\
        & & mAP & mAP \\
    \hline
        Faster & ResNet-50 & 77.8 & 42.2 \\
        Faster & ResNet-50-DCN & 80.5 & 42.1 \\
        Faster & ResNet-101 & 78.2 & 43.4 \\
        Faster & ResNet-101-DCN & 81.0 & 43.6 \\
        Cascade & ResNet-101 & 78.7 & 43.7 \\
        Cascade & ResNet-101-DCN & 81.6 & 43.9 \\
    \hline
    \end{tabular}
    \caption{Experiment results of baseline detectors on \textbf{Aluminum} and \textbf{Fabric} datasets.}
    \label{tab:baseline}
\end{table}

\begin{table}[t]
\small
    \centering
    \begin{tabular}{c|c|c|c}
    \hline
      Detector & Backbone & Processing & \textbf{Fabric} mAP\\
    \hline
      Faster & ResNet-50 & - & 42.2 \\
      Faster & ResNet-50 & Pre & 48.7 \\
      Faster & ResNet-50 & Inter & 48.0 \\
      Faster & ResNet-50 & Post & 47.2 \\
      Faster & ResNet-50 & Pre* & 45.6 \\
      Faster & ResNet-50 & Inter* & 47.8 \\
      Faster & ResNet-50 & Post* & 46.9 \\
    \hline
    \end{tabular}
    \caption{Experiment results of template reference processing with well-aligned templates on \textbf{Fabric} dataset. * denotes trained on shifted dataset.}
    \label{tab:tg1}
\vspace{-3mm}
\end{table}

\begin{table}[t]
\small
    \centering
    \begin{tabular}{c|c|c|c|c}
    \hline
      Detector & Backbone & TR & \textbf{Aluminum} & \textbf{Fabric}\\
      & & & mAP & mAP \\
    \hline
      Faster & ResNet-50 & - & 77.8 & 42.2 \\
      Faster & ResNet-50 & templates & - & 48.0 \\
      Faster & ResNet-50 & PTG & 79.2 & 46.4 \\
      Faster & ResNet-101 & - & 78.2 & 43.4 \\
      Faster & ResNet-101 & templates & - & 49.1 \\
      Faster & ResNet-101 & PTG & 79.7 & 47.0 \\
      Cascade & ResNet-101 & - & 78.7 & 43.7 \\
      Cascade & ResNet-101 & templates & - & 49.2 \\
      Cascade & ResNet-101 & PTG & 80.1 & 47.2 \\
    \hline
    \end{tabular}
    \caption{Experiment results of TR-Inter on \textbf{Aluminum} and \textbf{Fabric} datasets.}
    \label{tab:tg2}
\end{table}

\subsection{Performance of Baseline}
We adopt ResNet-50 and ResNet-101 as backbones and evaluate Faster R-CNN and Cascade R-CNN two architecture. We also try to equip the CNN backbone with deformable convolution, denoted as ResNet-50-DCN and ResNet-101-DCN in short. The performance of the baseline on \textbf{Aluminum} and \textbf{Fabric} datasets can be found in Table~\ref{tab:baseline}.

The third column shows the performance on the \textbf{Aluminum} dataset. We can find that replacing the backbone from ResNet-50 to ResNet-101 can bring $0.4$ mAP improvement. The Cascade R-CNN architecture can bring about $0.5$ mAP improvement, due to its powerful boundary regression ability. Deformable convolution can boost the detection performance a lot. We surprisingly find that deformable convolution can improve the mAP by $1.3$ on the ResNet-50 backbone, while the improvement on the ResNet-101 backbone can achieve $1.8$. We think this is because ResNet-101 has more deformable convolutional layers than ResNet-50, and thus can leed to more improvement.

For \textbf{Fabric} dataset, We can find that replacing the backbone from ResNet-50 to ResNet-101 can achieve $1.2$ mAP improvement from the fourth column. The Cascade R-CNN architecture can slightly improve $0.3$ mAP. The deformable convolution seems to perform not well on the fabric defect detection task.

\begin{table}[t]
\small
    \centering
    \begin{tabular}{c|c|c|c|c}
    \hline
        Detector & Backbone & Context & \textbf{Aluminum} & \textbf{Fabric}\\
        & & & mAP & mAP \\
    \hline
        Faster & ResNet-50 & - & 77.8 & 42.2 \\
        Faster & ResNet-101 & - & 78.2 & 43.4 \\
        Faster & ResNet-50 & \cite{he2016deep} & 77.9 & 42.0 \\
			Faster & ResNet-50 & CR(k=1.3) & 78.4 & 46.3\\
			Faster & ResNet-50 & CR(k=1.4) & 78.9 & 46.8\\
        Faster & ResNet-50 & CR(k=1.5) & 79.2 & 46.8\\
			Faster & ResNet-50 & CR(k=1.6) & 78.5 & 46.4\\
			Faster & ResNet-50 & CR(k=1.7) & 78.1 & 45.8\\
			Faster & ResNet-50 & CR*(k=1.5) & 77.3 & 44.8\\
        Faster & ResNet-101 & CR(k=1.5) & 79.5 & 47.3\\
    \hline
    \end{tabular}
    \caption{Experiment results of context reference.* denotes only using contextual region features for classification and bounding box regression.}
    \label{tab:cff}
\vspace{-3mm}
\end{table}

\begin{table*}[t]
\small
    \centering
    \begin{tabular}{c|c|c|c|c|c}
    \hline
    Detector & Backbone & Template Reference & Context Reference & \textbf{Aluminum} mAP & \textbf{Fabric} mAP\\
    \hline
        Faster & ResNet-50 & Pre & $\surd$ & - & 50.2 \\
        Faster & ResNet-50 & PTG & $\surd$ & 80.4 & 48.9 \\
        Faster & ResNet-50-DCN & Pre & $\surd$ & - & 50.3 \\
        Faster & ResNet-50-DCN & PTG & $\surd$ & 82.9 & 49.1 \\
        Faster & ResNet-101 & Pre & $\surd$ & - & 51.0 \\
        Faster & ResNet-101 & PTG & $\surd$ & 80.9 & 50.7 \\
        Faster & ResNet-101-DCN & Pre & $\surd$ & - & 51.2 \\
        Faster & ResNet-101-DCN & PTG & $\surd$ & 83.5 & 50.7 \\
        Cascade & ResNet-101 & Pre & $\surd$ & - & 51.3\\
        Cascade & ResNet-101 & PTG & $\surd$ & 81.2 & 50.8 \\
        Cascade & ResNet-101-DCN & Pre & $\surd$ & - & \textbf{51.4} \\
        Cascade & ResNet-101-DCN & PTG & $\surd$ & \textbf{83.9} & 50.9 \\
    \hline
    \end{tabular}
    \caption{Experiment results of reference-based defect detection networks on \textbf{Alunimum} and \textbf{Fabric} datasets.}
    \label{tab:all}
\vspace{-5mm}
\end{table*}

\subsection{Influence of Template Reference}

\begin{figure}[t]
    \centering
    \includegraphics[width=\linewidth]{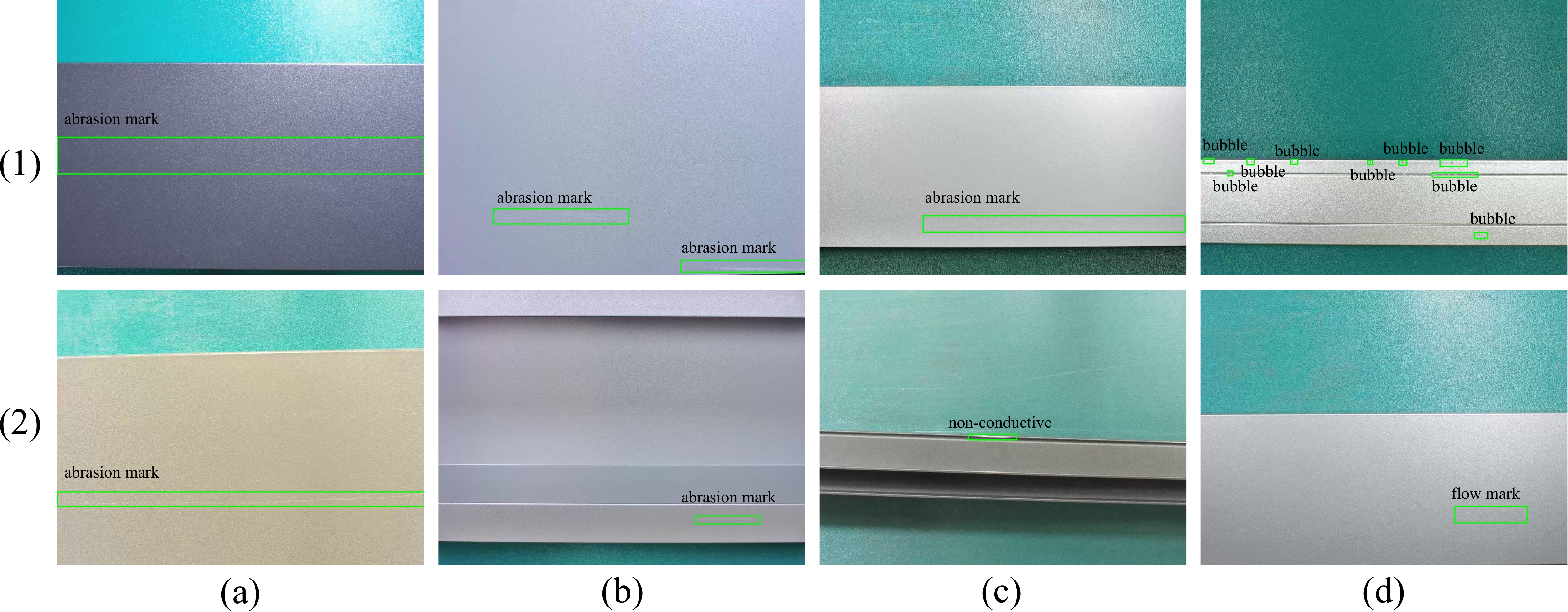}
    \caption{Detection cases of \textbf{Aluminum} dataset. The green boxes indicate the detected boxes. [Best zoom in for better view]}
    \label{fig:case1}
\end{figure}
\begin{figure}[t]
    \centering
    \includegraphics[width=\linewidth]{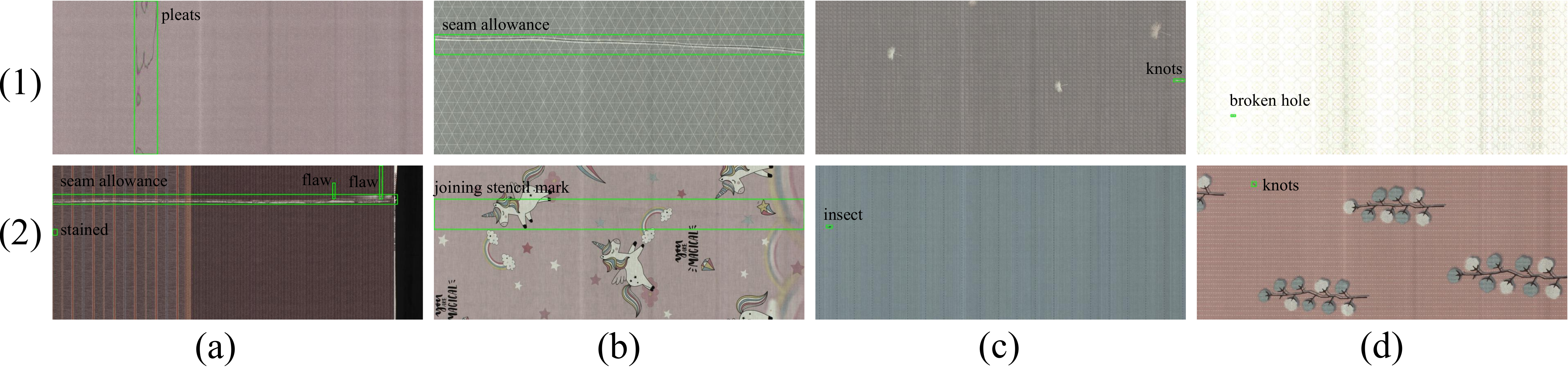}
    \caption{Detection cases of \textbf{Fabric} dataset. The green boxes indicate the detected boxes. [Best zoom in for better view]}
    \label{fig:case2}
\end{figure}

We first leverage the well-aligned templates images provided in \textbf{Fabric} dataset to evaluate the effectiveness of the three proposed template reference processing strategies. From Table~\ref{tab:tg1} we can find that TR-Pre, TR-Inter, and TR-Post can surprisingly improve the performance by $6.5$, $5.8$, $5.0$ mAP, respectively. Among these, TR-Pre performs the best and costs the least time because it only needs to forward one image to the network once. However, TR-Pre might be easily affected by pixel-gap. We randomly shift all images by $5-10$ pixels in training and testing sets to simulate the pixel-gap application scenario and evaluate the three processing methods again. We find that the performance of TR-Pre drops to $2.4$ mAP, while TR-Inter and TR-Post are more stable, whose performances only drop about $0.3$. Therefore, TR-Pre is the best choice in real-world applications if well-aligned template images are available, otherwise, TR-Inter and TR-Post might be better.

When well-aligned template images are unavailable, we can adopt a pseudo template generator (PTG) to generate pseudo templates and take the place of the real ones. We evaluate the effectiveness of PTG on both \textbf{Aluminum} and \textbf{Fabric} datasets, and report the experiments results in Table~\ref{tab:tg2}. We adopt TR-Inter as a template reference processing strategy and also compare with TR-Inter with real well-aligned templates if it is available. From Table~\ref{tab:tg2} we can find that PTG can significantly boost about $1.4$ mAP and $3.5$ mAP on \textbf{Aluminum} and \textbf{Fabric} datasets, respectively. On \textbf{Fabric} dataset, although there is still a performance gap between using PTG and real templates, PTG still can bring promising improvement.

\subsection{Influence of Context Reference}
Contextual reference enables each region proposal to involve surrounding information to perform classification and regression, so that can solve the issue caused by partial visual confusion. We report the performance of context reference on two datasets in Table~\ref{tab:cff}. From which we can find that, with both ResNet-50 and ResNet-101 backbone, context reference can steadily boost about $1.4$ mAP and $3.9$ mAP on \textbf{Aluminum} and \textbf{Fabric} datasets, respectively. We compare the effectiveness of difference coefficient values $k$, and find that $k=1.5$ performs the best. We also try to only use the feature of contextual regions $F^I_c$ to perform classification and bounding box regression, and we find that the performance drops a bit, which proves that the original region feature $F^I$ is necessary  for defect detection.

Utilizing contextual information has been proposed in \cite{he2016deep}. They consider the whole image as a global ROI and fuse contextual feature by adding the global region feature to all region proposals. Such a trick is shown effective on the MS-COCO benchmark. We re-implement this trick and apply it on both datasets, and also report such experiment results in Table~\ref{tab:cff}. We find that such a trick can not improve, or even drop the performance on both datasets. The reason might be the global is too coarse because of the down-sampling operation in the ROI wrapping layer, and such a coarse global feature can not provide additional useful information for region classification and regression.

\subsection{Performance of Reference-Based Defect Detector}
The template reference and context reference can work well together, and the two reference components construct the reference-based defect detection networks. Table~\ref{tab:all} shows the overall performance of the RDDN under different architecture and backbone settings. We can find that the combination of template reference and context reference performs well on all baseline detectors, and can bring at least $2.3$ and $7.3$ mAP improvement on \textbf{Aluminum} and \textbf{Fabric} datasets, respectively. We finally achieve $83.9$ mAP and $51.4$ mAP on the two datasets, respectively, which shows that our proposed reference-based defect detection network significantly outperforms vanilla object detectors.

\begin{table}[t]
\small
    \centering
    \begin{tabular}{c|c|c}
\hline
Module & Memory (Paras / Feature) & Time Cost \\
\hline
ResNet-50 & 179.35M / 1454.88M & 11.2ms \\
FPN & 2.55M / 112.06M & 1.3ms \\
PTG & 367.60M / 305.13M & 6.9ms \\
RPN & 4.53M / 0.19M & 16.9ms \\
R-CNN & 106.34M / 0.11M & 8.3ms \\
\hline

    \end{tabular}
    \caption{Memory and time cost of modules in RDDN.}
    \label{tab:memory_time}
\vspace{-3mm}
\end{table}

\subsection{Memory Cost and Inference Time}
RDDN involves reference to assist the detection, thus the memory cost and inference time cost will increase. We report the memory cost and time cost of each module in Table \ref{tab:memory_time}. Since CR will only slightly affect the R-CNN module, we only analyze the memory and time cost of TR. 

In the training stage, a Faster R-CNN detector costs about 3.3 GB memory for parameters, features and optimizers. TR-Pre hardly increase the memory cost, because it is applied in data-level. TR-Inter will double the feature memory cost of CNN backbone, and thus the total memory cost will increase to about 6 GB. TR-Post will affect the feature memory cost in R-CNN, and can be ignore since the R-CNN module only costs 0.11M feature memory. PTG will bring about 20\% additional memory cost. The memory cost of TR is acceptable since it can bring considerable performance gain. The whole model can be easily trained on 12GB-GPU, or even 6GB-GPU if using mix-precision training.

For model inference, the backbone and the RPN take the most of the time cost. TR-Inter will double the time cost of backbone, which will bring about 30\% additional time cost. TR-Post will double the time cost of the whole detector, which will bring about 55\% additional time cost. PTG costs about 6.9ms, which is only 18\% of the total inference time.

\subsection{Visualization and Case Study}
We visualize some detection cases of \textbf{Aluminum} and \textbf{Fabric} datasets in Fig.\ref{fig:case1} and Fig.\ref{fig:case2}, respectively. All cases are detected by the detector with Faster R-CNN architecture and ResNet-101 backbone. We integrate PTG and context reference when training detectors on both datasets. When performing visualization, we only keep the detected boxes with confidence scores higher than $0.8$.

From Fig.\ref{fig:case1} we can find that with the help of template reference, the difference between foreground and background can be enhanced, and thus the defects can be easier to be detected, as the case (1)(b) and (1)(c). The case(1)(d) the partial visual confusion issue, where a small bubble may be possibly a complete defect or a partial defect. In \textbf{Aluminum} dataset, if several bubbles are close to each other, they should be grouped together and be considered as one defect. For this case, we can find that our detection result can effectively distinguish the difference. This demonstrates the effectiveness of context reference.

One interesting finding is that the textures cases in Fig.~\ref{fig:case2} do not exist in the training set. We can find that our proposed detector is robust to unseen texture even without the help of real well-aligned templates.

\section{Conclusion}
We analyze the characteristics and difficulties of defect data in the wild, and we propose a reference-based defect detection network (RDDN) by fully utilizing the data characteristics to solve the difficulties. To solve the texture shift (TS) issue, we propose to utilize template reference to reduce the texture information in image, feature, or region level, and make the network focus more on the defect area. The template reference can be a well-aligned template or the outputs of our designed pseudo template generator, which is in encoder-decoder like architecture and can be jointly trained with the detectors. We leverage context reference to overcome the difficulty caused by partial visual confusion (PVC). The RDDN has rich application scenarios in the wild, and can significantly improve the defect detection performance.





\ifCLASSOPTIONcaptionsoff
  \newpage
\fi



\bibliographystyle{IEEEtran}
\bibliography{ref}
%

%
\vspace{-13mm}
\begin{IEEEbiography}[{\includegraphics[width=1in,height=1.25in,clip,keepaspectratio]{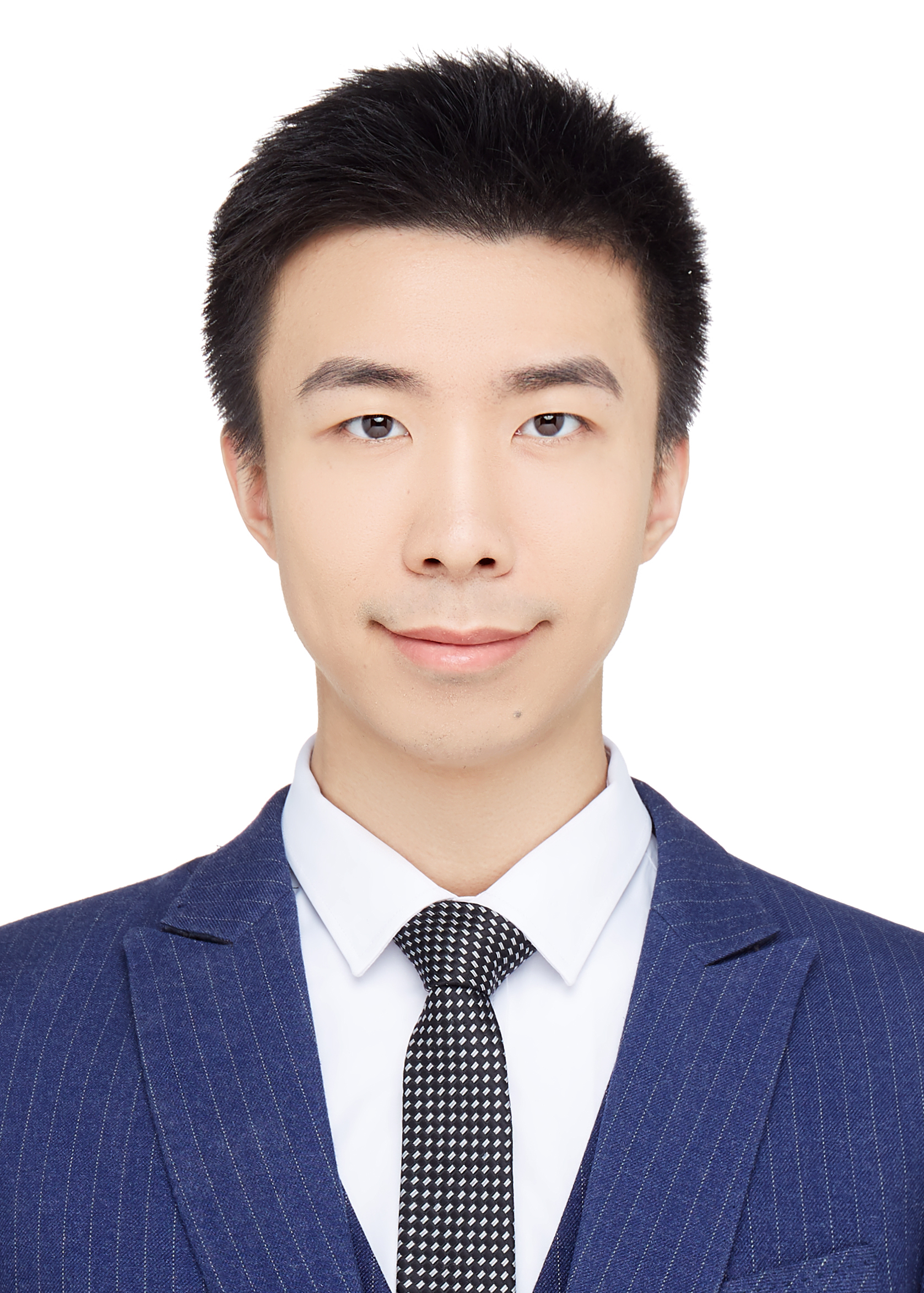}}]{Zhaoyang Zeng}
received the B.S. degree from the School of Data and Computer Science, Sun Yat-sen University, Guangzhou, China, in 2016, where he is currently pursuing the Ph.D. degree. His research interests
include object detection and vision-language understanding.
\end{IEEEbiography}
\vspace{-13 mm}

\begin{IEEEbiography}[{\includegraphics[width=1in,height=1.25in,clip,keepaspectratio]{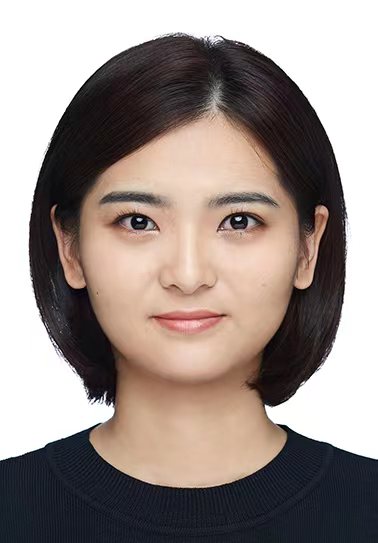}}]{Bei Liu}
is a researcher in Multimedia Search and Mining Group, Microsoft Research Asia (MSRA). Before joining Microsoft, she received her Ph.D. and master degree from Department of Social Informatics, Kyoto University, Jappan, in 2018 and 2014, respectively. She received the B.S. degree in the Institute of Software, Nanjing University, China, in 2011. Her current research interests include Vision and Language, visual creation and object detection. She received the Best Paper Award of ACM Multimedia 2018. She serves as reviewers for IEEE Trans. on Multimedia, ACM Multimedia, IEEE International Conference on Multimedia and Expo (ICME), ACM International Conference on Multimedia Retrieval (ICMR). She also receieved IEEE Trans. on Multimedia 2020 Outstanding Reviewer Award.
\end{IEEEbiography}
\vspace{-13 mm}

\begin{IEEEbiography}[{\includegraphics[width=1in,height=1.25in,clip,keepaspectratio]{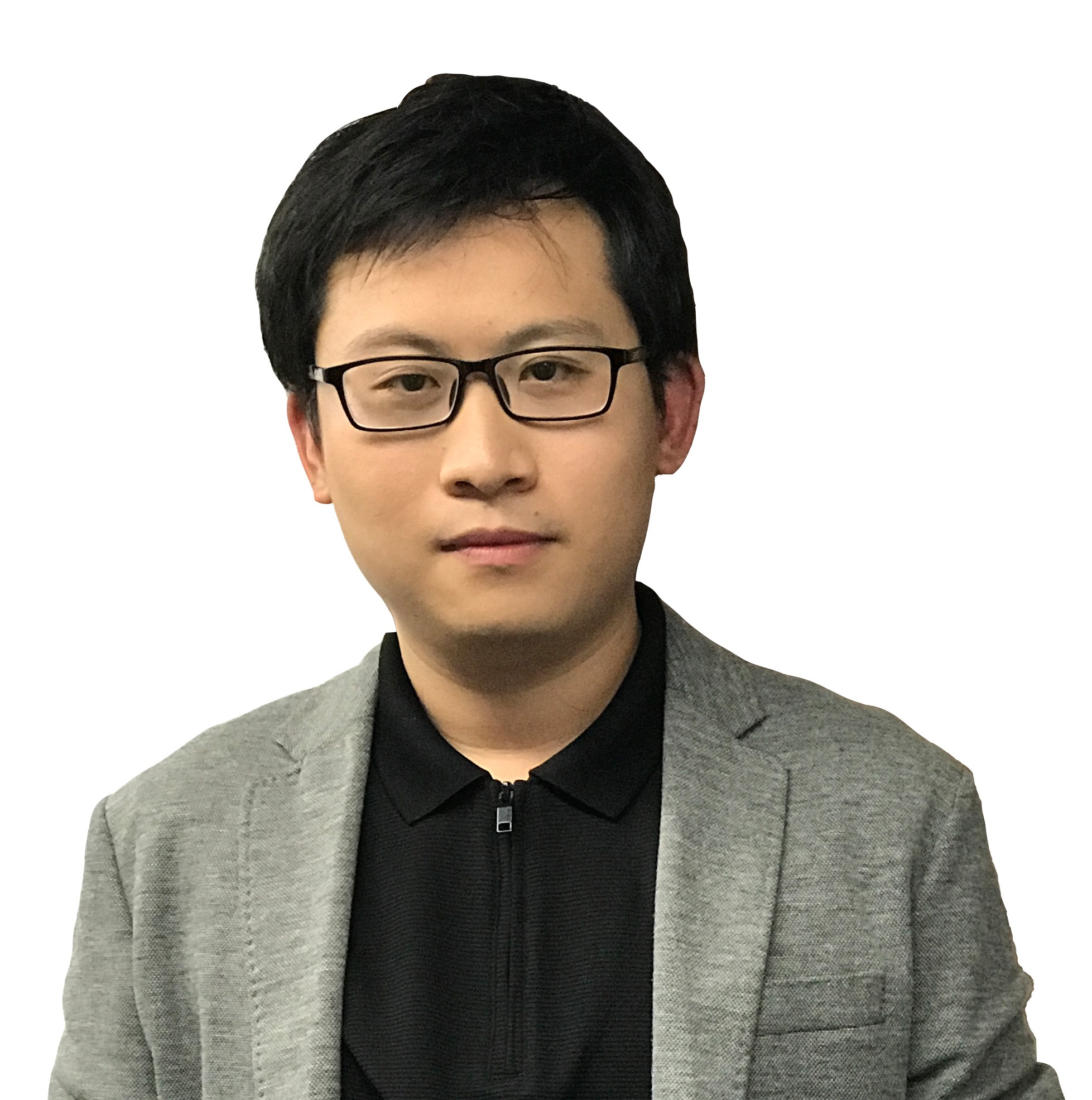}}]{Jianlong Fu}
is a Senior Research Manager with the Multimedia Search and Mining Group, Microsoft Research Asia (MSRA). He received his Ph.D. degree from the Institute of Automation, Chinese Academy of Science in 2015. His research interests include computer vision, and multimedia content understanding. He has authored or coauthored over 80 publications in journals and conferences, and 1 book chapter. He is an area chair of ACM Multimedia 2018, 2021, ICME 2019, 2020. He is a recipient of 2018 ACM Multimedia Best Paper Award.
\end{IEEEbiography}
\vspace{-13 mm}

\begin{IEEEbiography}[{\includegraphics[width=1in,height=1.25in,clip,keepaspectratio]{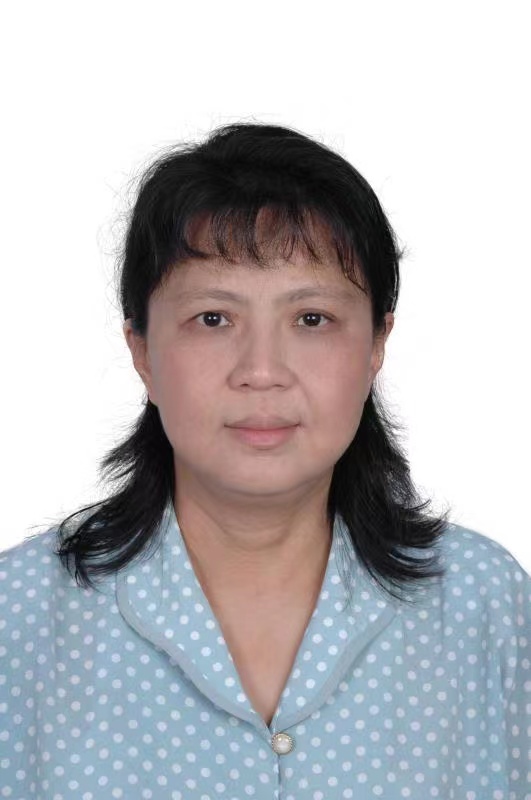}}]{Hongyang Chao}
received the B.S. and Ph.D. degrees in computational mathematics from Sun Yet-sen University, Guangzhou, China. In 1988, she joined the Department of Computer Science, Sun Yet-sen University, where she was initially an Assistant Professor and later became an Associate Professor. She is currently a Full Professor with the School of Data and Computer Science. She has published extensively in the area of image/video processing and holds 3 U.S. patents and four Chinese patents in the related area. Her current research interests include image and video processing, image and video compression, massive multimedia data analysis, and content-based image (video) retrieval.
\end{IEEEbiography}







\end{document}